\documentclass{article}

    \PassOptionsToPackage{numbers, compress}{natbib}

\usepackage[preprint]{neurips_2023}

\usepackage[utf8]{inputenc} 
\usepackage[T1]{fontenc}    
\usepackage{hyperref}       
\usepackage{url}            
\usepackage{booktabs}       
\usepackage{amsfonts}       
\usepackage{nicefrac}       
\usepackage{microtype}      
\usepackage{xcolor}         

\usepackage{amsmath,amsfonts,amsthm}
\usepackage{amsmath,bm}
\usepackage{amssymb}
\usepackage{subfigure}
\usepackage{graphicx}
\usepackage{enumitem} 
\usepackage{algorithmic,algorithm}
\usepackage{multirow}

\usepackage[utf8]{inputenc} 
\usepackage[T1]{fontenc}    
\usepackage{hyperref}       
\usepackage{url}            
\usepackage{booktabs}       
\usepackage{amsfonts}       
\usepackage{nicefrac}       
\usepackage{microtype}      
\usepackage{xcolor}         
\usepackage{enumitem}

\newcommand{\adv}{\mathrm{adv}}

\newcommand{\Appendix}[1]{the full version for}

\newcommand{\bdelta}{\bm{\delta}}

\renewcommand{\c}{\mathbf{c}}

\newcommand{\x}{\bm{x}}

\newcommand{\C}{\mathbf{C}}

\newcommand{\red}[1]{{\color{red}#1}}

\newcommand{\cL}{\mathcal{L}}

\title{Shielding the Unseen: Privacy Protection through Poisoning NeRF with Spatial Deformation}

\author{%
  Yihan Wu \\
  University of Maryland\\
  \texttt{ywu42@umd.edu} \\
  \And
    Brandon Y. Feng \\
  University of Maryland\\
  \texttt{brandon.fengys@gmail.com} \\
  \And
   Heng Huang \\
  University of Maryland\\
  \texttt{heng@umd.edu} \\
}

\begin{document}

\maketitle

\begin{abstract}
In this paper, we introduce an innovative method of safeguarding user privacy against the generative capabilities of Neural Radiance Fields (NeRF) models. Our novel poisoning attack method induces changes to observed views that are imperceptible to the human eye, yet potent enough to disrupt NeRF's ability to accurately reconstruct a 3D scene. To achieve this, we devise a bi-level optimization algorithm incorporating a Projected Gradient Descent (PGD)-based spatial deformation.
We extensively test our approach on two common NeRF benchmark datasets consisting of 29 real-world scenes with high-quality images. Our results compellingly demonstrate that our privacy-preserving method significantly impairs NeRF's performance across these benchmark datasets. Additionally, we show that our method is adaptable and versatile, functioning across various perturbation strengths and NeRF architectures.
This work offers valuable insights into NeRF's vulnerabilities and emphasizes the need to account for such potential privacy risks when developing robust 3D scene reconstruction algorithms. Our study contributes to the larger conversation surrounding responsible AI and generative machine learning, aiming to protect user privacy and respect creative ownership in the digital age.
\end{abstract}

\section{Introduction}
Neural Radiance Fields (NeRF) represents a significant milestone in novel view synthesis, propelling progress in 3D computer vision and graphics, and various applications ranging from extended reality to 3D generative AI, e-commerce, and robotics. However, these promising developments coincide with the ongoing urgent dialogue surrounding the responsible use of generative machine learning models for photo-realistic content generation. Despite the growing recognition of NeRF's potential, the research community has not yet thoroughly engaged with the crucial discussions surrounding the responsible deployment and distribution of NeRF-generated content.

The ease with which NeRF can create high-quality 3D content from internet images is a double-edged sword. With the abundance of user-uploaded images on social media and internet platforms, an unauthorized party could easily generate a 3D reconstruction of a user's surroundings {\it without consent}, even inferring private and personally identifiable information~\citep{mcamiswriting} from these immersive renderings.
As we anticipate further progress in NeRF's ability to synthesize novel views, even from internet images in-the-wild~\citep{MartinBrualla2020NeRFIT}, we must confront important questions: What if users don't want these novel views synthesized from their uploaded images? Do they have the right to refuse? How can we, as machine learning researchers, equip them with a protective shield guarding their own images?

Motivated by these potential privacy and security implications, we aim to fuse recent advancements in adversarial machine learning with the emerging issue of user privacy protection against the generative power of NeRF. Specifically, we present a method to introduce imperceptible alterations to 2D images that thwart view synthesis via NeRF. This approach essentially "vaccinates" user-captured images against unauthorized 3D scene generation through NeRF while preserving the original visual qualities of the 2D images for traditional viewing purposes like photo sharing.

To achieve our goal, we turn the strength of NeRF into its weakness - the ability of NeRF to capture local high-frequency details for photo-realistic rendering actually makes it sensitive to input variations. This vulnerability is rooted in the optimization of NeRF, where the pixel-wise L2 reconstruction loss leads to overfitting, especially when no prior information about the scene or object is available to regularize the NeRF model. We propose to exploit this vulnerability by introducing perturbations to the input images, leading to visible artifacts in the reconstructed scene. In essence, we manipulate NeRF's own precision against it, impairing its ability to generate accurate renderings by subtly distorting the data that it relies upon.

Several questions arise: How do we design and apply these perturbations? How much alteration is necessary to impact NeRF's robustness? Can we use gradient-based methods to derive these perturbations? While gradient-based approaches such as the PGD attack~\citep{madry2017towards} are popular, they are limited to altering pixel colors within a certain range and cannot distort images spatially, making it challenging to interfere with NeRF. We propose a solution in the form of spatial deformation~\citep{xiaospatially}, which distorts images spatially and effectively compromises NeRF's 3D space modeling.

A critical challenge lies in the necessity of executing these attacks during training time, as NeRF utilizes the provided images as training data, unlike traditional inference-time attacks on an already-trained model. We discuss how to conduct training-time attacks and generate imperceptible image alterations capable of disrupting NeRF. In this paper, we describe a poisoning recipe to generate image perturbations that result in an essentially "un-trainable" NeRF.

\textbf{Our contributions:} 
Our work's significance lies in safeguarding user privacy by preventing unauthorized leaks and intrusion of their 3D surrounding environment. 
\begin{itemize}
    \item We present the first method for poisoning NeRF using spatial deformation and develop a bi-level algorithm employing a PGD-based attack to create training-time perturbations on NeRF.
    \item We generate a human-imperceptible poisoned dataset including $\sim30$ real forward facing scenes to evaluate the effectiveness of our approach. Our findings demonstrate that this poisoned dataset significantly impairs the rendering quality of NeRF. Further, our method displays both adaptability and versatility, proving effective across a range of perturbation strengths and distinct NeRF architectures.
\item Our research represents an initial step towards protecting user privacy in NeRF models by introducing an imperceptible spatial deformation that renders NeRF training ineffective. In this regard, our research bridges the gap in the literature by exploring NeRF's potential vulnerabilities to poisoning attacks. Our findings also open new avenues for research in the field of adversarial machine learning.
\end{itemize}

In the broader context of the ongoing dialogue around responsible AI and generative machine learning models, this work seeks to contribute to a more ethical and controlled use of these powerful technologies. By recognizing and addressing the potential for misuse, we hope to encourage the development of safeguards that protect user privacy and respect content ownership, even as we continue to advance in our ability to generate and manipulate visual content with machine learning.

\section{Related Works}
\textbf{Novel view synthesis.}
 View synthesis techniques enabling the creation of photo-realistic novel views of a scene from a limited number of images. One approach to scene geometry construction is to use local warps or by explicitly computing the camera poses and 3D point cloud of the scene \citep{buehler2001unstructured,chaurasia2013depth}. Another approach involves using light field rendering, which allows for the implicit reconstruction of the scene geometry and the creation of photorealistic novel views \citep{gortler1996lumigraph,levoy1996light}. Proxy scene geometry enhancement techniques aim to improve the quality of the reconstructed 3D model. \citep{penner2017soft} used soft 3D reconstruction, which produces a smooth and continuous scene representation by incorporating learned deep features. Learning-based dense depth maps can also be used to improve the quality of the scene geometry \citep{flynn2016deepstereo}, as well as multiplane images (MPIs) \citep{choi2019extreme,flynn2019deepview,srinivasan2019pushing}. Additional deep features can also be learned to enhance the quality of the reconstructed scene \citep{hedman2018deep,riegler2020free}. Voxel-based implicit scene representations are another approach to proxy geometry enhancement \citep{sitzmann2019scene}.

Neural Radiance Field (NeRF) is a state-of-the-art technique for view synthesis. NeRF directly represents the scene as a function that maps spatial coordinates and viewing angles to local point radiance \citep{barron2022mip,mildenhall2021nerf,sitzmann2019scene}. By using differentiable volumetric rendering techniques, NeRF is able to generate novel views of the scene from arbitrary viewpoints \citep{kajiya1984ray,mildenhall2021nerf}. Moreover, NeRF achieves photo-realistic novel view synthesis with only RGB supervision and known camera poses, making it an appealing choice for many applications. Numerous extensions of NeRF have been proposed to address various limitations of the original approach. For instance, to accelerate the rendering process, several works have explored techniques such as multi-level sampling and hierarchical modeling \citep{reiser2021kilonerf,wang2022fourier}. To handle scenes with large scale variations, researchers have proposed multi-scale NeRF \citep{barron2022mip,tancik2022block}. For dynamic scenes, approaches such as neural radiance flow and neural event representations have been proposed \citep{li2021neural,park2021hypernerf,tretschk2021non}. Additionally, to improve the rendering of specular surfaces, specular NeRF has been introduced \citep{verbin2022ref}. Our work focuses on poisoning the static NeRF approaches.

\textbf{Poisoning attacks in deep learning.} 
The core idea of poisoning attacks is to introduce malicious data into training
datasets of target models to hinder the model training \citep{barreno2010security,barreno2006can}. 
Optimization-based data manipulation is one of the most popular poisoning methods. \citep{munoz2017towards} applied poisoning attacks to deep learning models using backgradient optimization, which allowed any sample to attack a designated class without the need for careful selection. Another notable method is MetaPoison \citep{huang2020metapoison}, which is a first-order optimization method that approximates the bilevel problem via meta-learning and constructs poisoning samples against deep learning models. \citep{shafahi2018poison} proposed a clean-labels poisoning attack against deep learning models. The poisoning objective function is designed to constrain the visual similarity between the poisoning sample and a benign sample while containing poisoning information. This objective function is optimized via a forward-backward-splitting iterative procedure \citep{goldstein2014field}.
The advantages of optimization-based poisoning attacks include the control of poisoning samples and the ability to accomplish various adversarial goals with a reasonable design. In our work, we design a bi-level optimization objective for poisoning NeRF.

\textbf{Robustness on NeRF.} 
To the best of our knowledge, there has been no investigation into poisoning attacks on Neural Radiance Fields (NeRF). The most closely related work, albeit orthogonal, is by \citep{wang2023benchmarking}, which conducted a benchmarking study to measure the robustness of NeRF under specific non-optimizable image corruptions, such as defocus blur, fog, and Gaussian noise. Our work, however, differs from their approach in that we introduce human-imperceptible spatial perturbations to the training images. These perturbations are obtained through a bi-level optimization process, which allows us to explore the effects of such perturbations on the performance of NeRF models.

\begin{figure}[t]
    \centering
    \includegraphics[height=5.5cm]{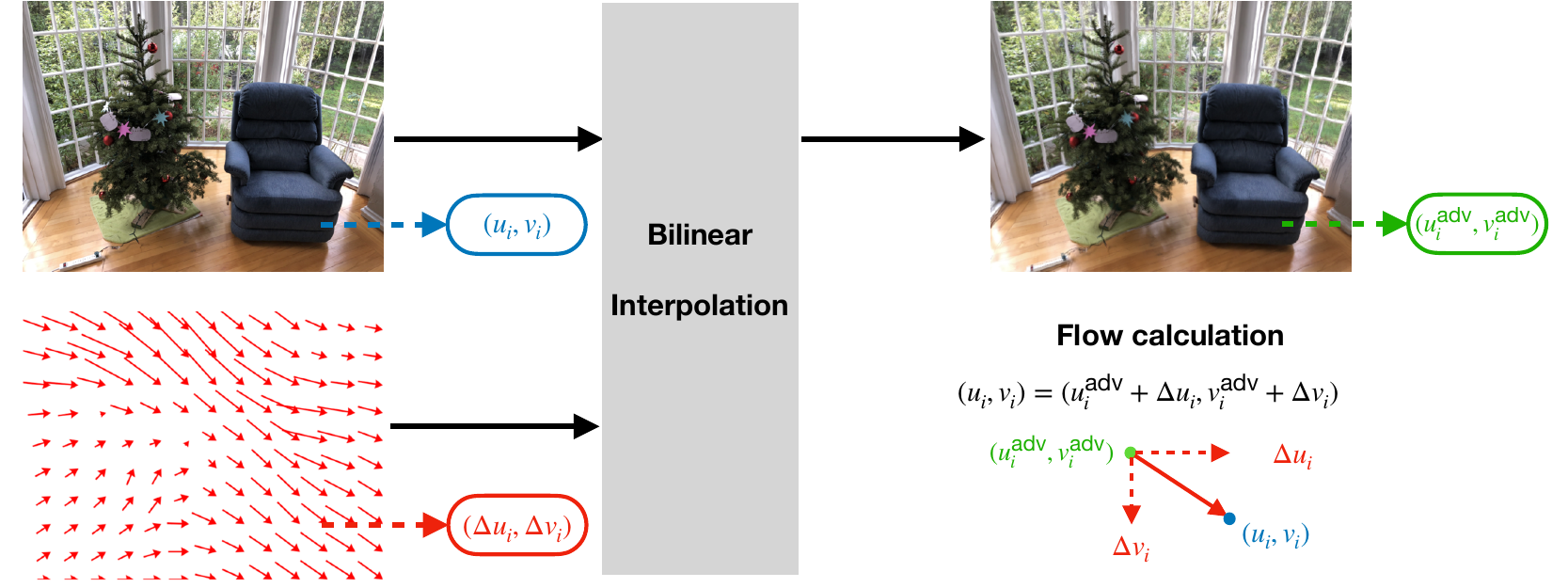}
    \vspace{-0.4cm}
    \caption{Creating adversarial examples using spatial transformation: The blue point represents the coordinate of a pixel in the output adversarial image, while the green point corresponds to its respective pixel in the input image. The red flow field indicates the displacement from pixels in the adversarial image to the corresponding pixels in the input image.}
    \label{fig:results}
    \vspace{-0.4cm}
\end{figure}
\section{Methods}
In this section, we present our bi-level optimization algorithm, which employs Projected Gradient Descent (PGD) with spatial deformation, designed to effectively execute poisoning attacks on Neural Radiance Fields (NeRF).

\noindent \textbf{Goal.} Our aim is to introduce human-imperceptible alterations to the observed views in a manner that causes the standard Neural Radiance Fields (NeRF) model to fail in accurately reconstructing the 3D scene from the given 2D observed images.

\noindent \textbf{Neural Radiance Fields.} View synthesis aims at constructing specific scene geometry from images captured at various positions and using local warps to synthesize high-quality novel views of a scene.
Neural Radiance Field (NeRF) is one of the SOTA view synthesis techniques, which utilizes implicit MLPs, represented by $\Theta$, to capture a static 3D scene. These MLPs map the 3D position $(x, y, z)$ and viewing direction $(\theta, \phi)$ to their corresponding color $\c$ and density $\sigma$:
$$(\c, \sigma) = \textrm{MLP}_{\Theta}(x, y, z, \theta, \phi)$$
The pixel color is computed by applying volume rendering to the ray $r$ emitted from the camera origin:
\begin{equation}\label{eq:nerfloss}
    \begin{split}
        \hat\C(r) &= \sum_{i=1}^N T(i)(1 - \exp(-\sigma(i)\delta(i)))\c(i),\\
        & T(i) = \exp(-\sum_{j=1}^{i-1}\sigma(j)\delta(j)),
    \end{split}
\end{equation}


where $\delta(i)$ represents the distance between two adjacent sample points along the ray, $N$ represents the count of samples along each ray. Furthermore, $T(i)$ refers to the accumulated transparency at sample $i$. Since the volume rendering process is differentiable, we can optimize the radiance fields by minimizing the reconstruction error between the rendered color $\hat\C$ and the actual color $\C$, which is expressed as:
$$\cL:=\sum_{r}||\hat\C(r) - \C(r)||_2$$

\begin{figure}
    \centering
    \includegraphics[height=6.8cm]{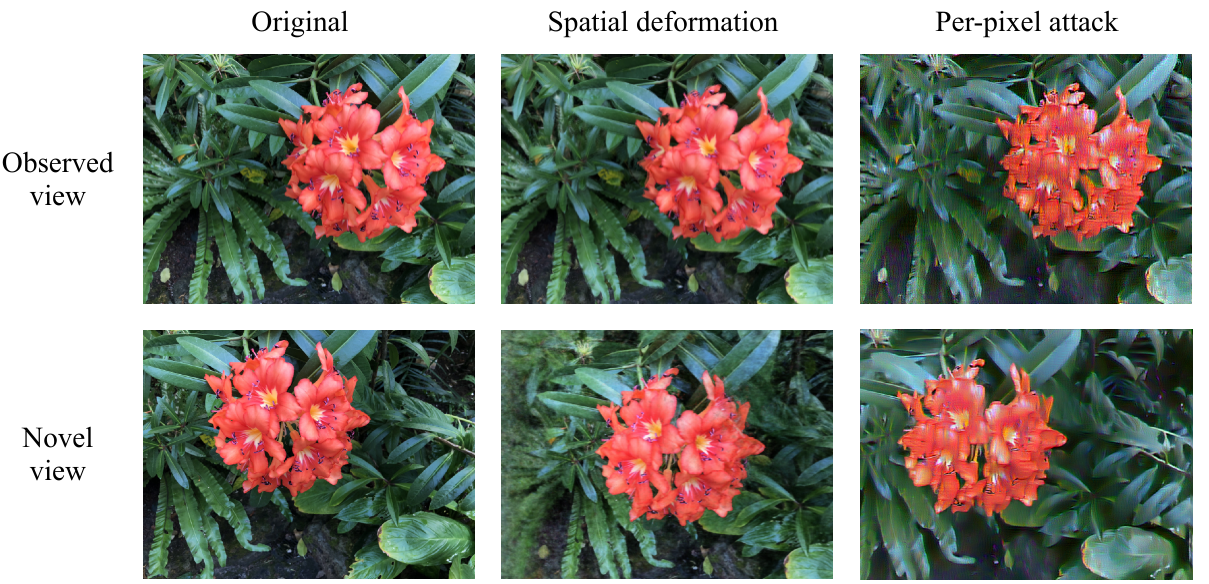}
    \vspace{-0.3cm}
    \caption{Poisoning NeRF using adversarial examples created through spatial deformation and per-pixel perturbations can be observed in the plot. It is evident that per-pixel perturbations considerably alter the color of the original images, which is reflected in the color change of the generated novel view. Meanwhile, spatial deformation introduces imperceptible modifications to the observed view, yet results in a blurred appearance in the generated view (as seen on the left side of the novel view).}
    \label{fig:pgdcompare}
    \vspace{-0.3cm}
\end{figure}
\noindent\textbf{Spatial deformation \citep{xiaospatially}.}
Most existing PGD-based adversarial perturbations directly modify pixel values, which may sometimes produce noticeable artifacts. Instead, spatial deformation smoothly change the geometry of the scene while keeping the original appearance, producing more perceptually realistic adversarial examples. 

When performing spatial transformation, the value of the i-th pixel and its 2D coordinate location $(u_i^\adv, v_i^\adv)$ in the adversarial image $\x^\adv$ are denoted as $\x_i^\adv$. It is assumed that $\x_i^\adv$ is transformed from the pixel $\x_i$ in the original image. Per-pixel flow field $f$ is used to synthesize the adversarial image $\x^\adv$ by utilizing pixels from the input $\x$. The optimization of displacement in each image dimension for the i-th pixel within $\x^\adv$ at the pixel location $(u_i^\adv, v_i^\adv)$ is carried out using the flow vector $f_i:= (\Delta u_i, \Delta v_i)$, which goes from a pixel $\x_i^\adv$ in the adversarial image to its corresponding pixel $\x_i$ in the input image. Therefore, the location of the corresponding pixel $\x_i$ can be obtained as $\x_i^\adv$ of the adversarial image $\x^\adv$ with the pixel value of input $\x$ from the location $(u_i, v_i) = (u_i^\adv + \Delta u_i, v_i^\adv + \Delta v_i)$. To transform the input image with the flow field, differentiable bilinear interpolation \citep{jaderberg2015spatial} is used, as the $(u_i, v_i)$ coordinates may be non-integer values and do not necessarily lie on the integer image grid. The value of $\x_i^\adv$ is calculated using the equation:
\begin{equation}\label{eq:spatial deform}
    \x_i^\adv = \sum_{\x_q\in N(u_i,v_i)} \x_q(1 - |u_i - u_q|)(1 - |v_i - v_q|),
\end{equation}
where $N(u_i, v_i)$ denotes the indices of the 4-pixel neighbors at the location $(u_i, v_i)$ (top-left, top-right, bottom-left, bottom-right). By applying \autoref{eq:spatial deform} to every pixel $\x_i^\adv$, the adversarial image $\x^\adv$ can be obtained. It is important to note that the adversarial image $\x^\adv$ is differentiable with respect to the flow field $f$, and the estimated flow field captures the amount of spatial transformation required to fool the classifier.

\begin{algorithm}[t]
\caption{Poisoning attack on NeRF}
\label{alg:1}
\textbf{Input}: Observed views $X$; model parameters $\Theta$;  with and without AWP; number of epochs $T$; perturbation strength $\rho$, learning rate $\alpha$ for updating $\Theta$, number of steps $k$ for updating $\Theta$, learning rate $\alpha'$ for updating perturbation $\bdelta$, total learning epochs $m$.
\begin{algorithmic}[1] 
\STATE Initialize model parameters $\Theta$, initialize spatial perturbation $\delta$.
\FOR{$t\in$ 1:m}
\FOR{$s\in$ 1:k}
\STATE Compute the loss: $\cL= \sum_r||\hat\C(r)-\C(r)||_2$ via \autoref{eq:nerfloss} on the perturbed set $\{\x\circ\bdelta\}_{\x\in X}$;\\
\STATE	Update the $\Theta$ through gradient descent: $\Theta = \Theta-\alpha\nabla_{\Theta}\cL;$\\
\ENDFOR
\STATE	Compute the gradient of $\bdelta$: $\bm{g}_{\bdelta} = \nabla_{\bdelta} \sum_{r}||\hat\C(r;\Theta) - \C(r;\{\x\}_{\x\in X})||_2$\\ 
\STATE	Update the perturbation via $\bdelta = \prod_{B(\bdelta,\rho)}( \bdelta+\alpha' \bm{g}_{\bdelta})$ 
\ENDFOR
\STATE \textbf{return} ${\bm{\theta}}_T$
\end{algorithmic}
\end{algorithm}
\noindent\textbf{Compare with per-pixel poisoning attack.} One might naturally question why we do not employ the popular PGD attack \citep{madry2017towards}, which directly modifies the pixel values of the image, for poisoning NeRF. There are two primary reasons for this choice: 1) NeRF relies on the spatial relationships of the observed views, so we believe that attacking the spatial aspects of the observed view would be more efficient for poisoning NeRF than altering pixel values. 2) As illustrated in \autoref{fig:pgdcompare}, the PGD attack significantly changes the color of the observed view, causing the generated novel views to exhibit different colors compared to the ground truth and thus resulting in low performance metrics. Consequently, it is difficult to determine whether the PGD attack has successfully poisoned the NeRF or if the low performance metric is simply a byproduct of the color change.

\noindent\textbf{Poisoning NeRF.} Diverging from numerous adversarial attacks that target machine learning models during the inference phase, our objective is to perturb the observed images at the training stage, causing the NeRF model to fail in learning the 3D representation from the provided 2D images. Our approach involves constructing a perturbation-parameterized classifier and optimizing the perturbation by maximizing the NeRF training loss on this classifier. Given an observed dataset $X$, a flow perturbation $\delta$, and a flow budget $\rho$, our optimization objective is formulated as the following bi-level optimization problem:
\begin{equation}\label{eq:opt}
    \begin{split}
        &\max_{||\bdelta||_\infty\leq\rho} \sum_{r}||\hat\C(r;\Theta^*(\{\x\circ\bdelta\}_{\x\in X})) - \C(r;\{\x\}_{\x\in X})||_2,\\
        &\textrm{subject to } \Theta^*(\{\x\circ\bdelta\}_{\x\in X}) = \min_{\Theta}\sum_{r}||\hat\C(r;\Theta) - \C(r;\{\x\circ\bdelta\}_{\x\in X})||_2,
    \end{split}
\end{equation}
where $\{\x\}_{\x\in X}$ refers the set of observed images, $\{\x\circ\bdelta\}_{\x\in X}$ represents the set of perturbed observed images, and $\C(r,\{\x\circ\bdelta\}_{\x\in X})$ is the ground truth color from $\{\x\circ\bdelta\}_{\x\in X}$. In the first level of our optimization problem $\Theta^*(\{\x\circ\bdelta\}_{\x\in X}) = \min_{\Theta}\sum_{r}||\hat\C(r;\Theta) - \C(r;\{\x\circ\bdelta\}_{\x\in X})||_2$, we build the perturbation-parameterized classifier by minimizing the NeRF loss on the perturbed training images. In the second level of our optimization problem $\max_{||\bdelta||_\infty\leq\rho} \sum_{r}||\hat\C(r;\Theta^*(\{\x\circ\bdelta\}_{\x\in X})) - \C(r;\{\x\}_{\x\in X})||_2$, we obtain the perturbations through maximizing the NeRF loss on the perturbation-parameterized classifier. In the following paragraph, we introduce the algorithm we designed for this bi-level optimization problem.


\noindent\textbf{Algorithm.} Directly addressing the aforementioned bi-level optimization problem is computationally challenging; therefore, we introduce an algorithm to learn the perturbations. To tackle the first level of the optimization problem in \autoref{eq:opt}, we employ multi-step gradient descent to approximate the optimal perturbation-parameterized classifier $\Theta^*(\{\x\circ\bdelta\}_{\x\in X})$ (refer to Step 1 below). For the second level of the optimization problem in \autoref{eq:opt}, we utilize one-step Projected Gradient Descent (PGD) to update the perturbation $\bdelta$ (see Step 2 below). The primary components of our algorithm can be summarized as follows:
\begin{itemize}
    \item Step 1: updating $\Theta$ by recursively learning on the perturbed dataset $\{\x\circ\bdelta\}_{\x\in X}$ for $k$ steps:
    $$\Theta \leftarrow \Theta-\alpha\nabla_{\Theta} \sum_{r}||\hat\C(r;\Theta) - \C(r;\{\x\circ\bdelta\}_{\x\in X})||_2.$$
    \item Step 2: updating $\bdelta$ by maximizing the rendering error with one-step PGD:
    $$\bdelta \leftarrow \prod_{B(\bdelta,\rho)}(\bdelta+\alpha'\nabla_{\bdelta} \sum_{r}||\hat\C(r;\Theta) - \C(r;\{\x\}_{\x\in X})||_2).$$
    
\end{itemize}
Since in Step 1, $\Theta$ is related to $\bdelta$, we can calculate the gradient of $\bdelta$ w.r.t. $\Theta$ in Step 2. We then repeatedly apply Step 1 and 2 for $m$ times to obtain the final perturbations. The detailed algorithm is in Alg.~\ref{alg:1}.

\begin{table}[t]
\centering
\caption{Our poisoning algorithm significantly reduces novel view synthesis performance on real-world datasets. We report PSNR/SSIM (higher is better) and LPIPS \citep{zhang2018unreasonable} (lower is better). The labels "Ori" and "Poi" denote the performance when trained with clean and perturbed views, respectively. \textbf{Top.} The average quantitative results for the 8 LLFF scenes and 21 Real-Forward-Facing (full) scenes are shown, with the red values in brackets indicating the reduction in performance.  \textbf{Bottom.}  Per-scene quantitative results on the specific LLFF scenes.}
\label{tab:general results}
\scalebox{0.93}{
\begin{tabular}{ccccccc}
\toprule
                                    & \multicolumn{3}{c}{\textbf{LLFF} (8 scenes)} & \multicolumn{3}{c}{\textbf{Real-Forward-Facing} (21 scenes)} \\ \midrule
                                    & PSNR$\uparrow$      & SSIM$\uparrow$      & LPIPS$\downarrow$     & PSNR$\uparrow$              & SSIM$\uparrow$             & LPIPS$\downarrow$          \\ \midrule
  Ori.  & 26.73     & 0.839     & 0.204     & 22.20             & 0.712            & 0.281            \\
                          Poi. & 19.99\red{(-6.74)}     & 0.539\red{(-0.300)}    & 0.526\red{(+0.322)}   & 18.65\red{(-3.55)}             & 0.417\red{(-0.295)}            & 0.568\red{(+0.287)}            \\ \bottomrule
\end{tabular}
}
\vspace{0.2cm}

\begin{tabular}{lccccccccc}
\toprule
                       &           & \textbf{Fern} & \textbf{Fortress} & \textbf{Leaves} & \textbf{Flower} & \textbf{Horns} & \textbf{Orchids} & \textbf{Room} & \textbf{T-Rex} \\ \midrule
\multirow{2}{*}{PSNR$\uparrow$}  & Ori.  & 25.27         & 31.36             & 21.30           & 28.60           & 28.14          & 19.87            & 32.35         & 26.97          \\
                       & Poi. & 19.08         & 23.70             & 14.77           & 20.84           & 21.88          & 16.91            & 21.64         & 21.15          \\ \midrule
\multirow{2}{*}{SSIM$\uparrow$}  & Ori.  & 0.814         & 0.897             & 0.752           & 0.871           & 0.877          & 0.649            & 0.952         & 0.900          \\
                       & Poi. & 0.457         & 0.735             & 0.279           & 0.510           & 0.628          & 0.340            & 0.731         & 0.630          \\ \midrule
\multirow{2}{*}{LPIPS$\downarrow$} & Ori.  & 0.237         & 0.148             & 0.217           & 0.169           & 0.196          & 0.278            & 0.167         & 0.221          \\
                       & Poi. & 0.587         & 0.373             & 0.589           & 0.572           & 0.494          & 0.590            & 0.518         & 0.486          \\ \bottomrule
\end{tabular}
\end{table}

\begin{figure}
    \centering
    \includegraphics[height=10.5cm]{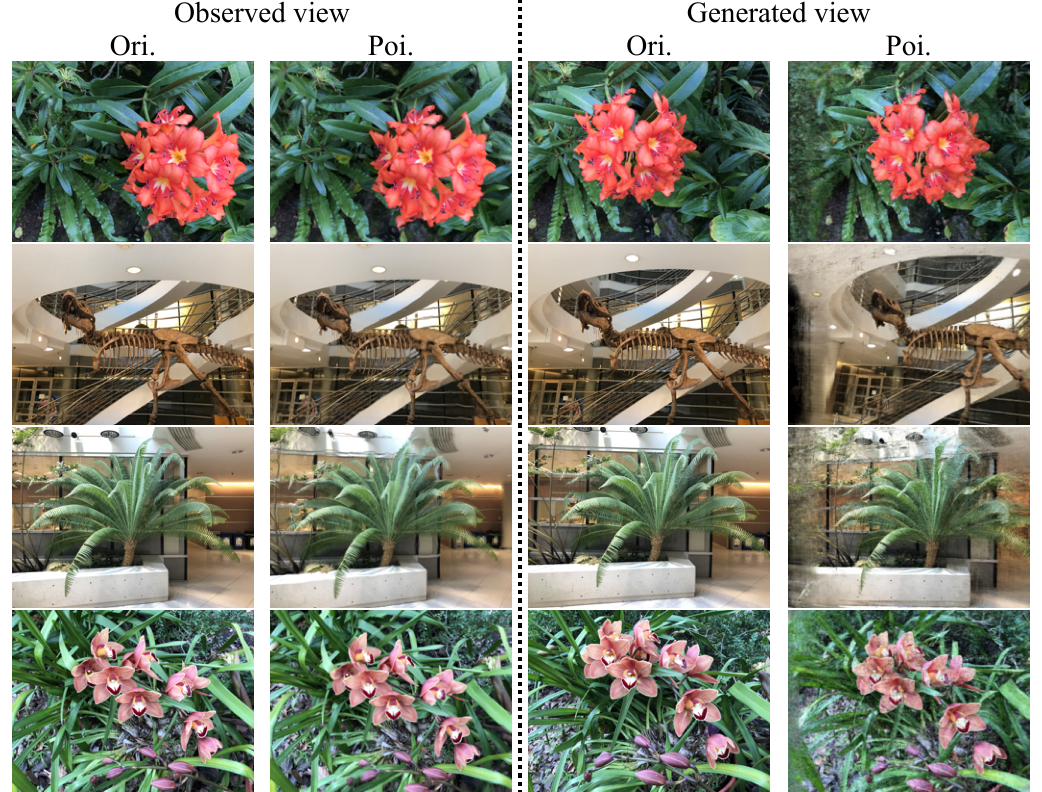}
    \caption{Comparisons on observed and generated views for the LLFF dataset. For the poisoning attack we apply spatial deformation with strength $\rho=10$. It is evident that the generated views are significantly impacted (particularly at the edges of the images) by subtle alterations in the observed views.}
    \label{fig:resultall}
    \vspace{-0.6cm}
\end{figure}

\section{Experiments}
In this section, we perform extensive experiments to assess the efficacy of our algorithms on NeRF models. We encourage the reader to watch our supplementary video, which effectively demonstrates the substantial reduction in rendering quality of novel views achieved by our poisoning method.

\textbf{Datasets.} 
Our experiments primarily focus on two datasets: the LLFF dataset and the Real-Forward-Facing (full) dataset. The LLFF dataset encompasses 8 real-world scenes, mainly composed of forward-facing images, while the Real-Forward-Facing dataset includes 21 real-world forward-facing scenes, offering a broader range of examples. During the data poisoning process, we initially introduce spatial deformations to the training images. Then we proceed to train NeRF on these perturbed views, which have been carefully designed to integrate the spatial deformations. By adopting this approach, we aim to thoroughly evaluate the effects of data poisoning on NeRF models across various real-world contexts.
.

\textbf{Models.} Given that our poisoning algorithm necessitates a substantial amount of gradient computation, it is essential to utilize an efficient NeRF backbone to minimize the running time. In our experiments, we mainly adopt TensoRF \citep{chen2022tensorf} for generating spatial perturbations. As a popular method to enable rapid NeRF training, TensoRF demonstrates both time and memory efficiency while maintaining strong performance. Furthermore, we demonstrate that the perturbations generated by poisoning TensoRF can be successfully transferred to the other NeRF backbones, highlighting the versatility of our approach.

\textbf{Settings.} 
For the TensoRF settings, we adhere to the original configurations employed for the LLFF dataset when generating perturbations. We utilize a small MLP consisting of two fully connected layers (with 128-channel hidden layers) and ReLU activation for neural features. The Adam optimizer \citep{kingma2014adam} is used, with initial learning rates of 0.02 for tensor factors and, when employing neural features, 0.001 for the MLP decoder. Our model is optimized using a batch size of 4096 pixel rays.
To achieve a coarse-to-fine reconstruction, we commence with an initial low-resolution grid consisting of $128\times128\times128$ voxels. Subsequently, we upsample the vectors and matrices linearly and bilinearly at steps 2000, 3000, 4000, 5500, and 7000, interpolating the number of voxels between $128\times128\times128$ and $640\times640\times640$ linearly in logarithmic space.

For the poisoning settings, we establish an $\ell_\infty$ budget of spatial deformation $\rho = 10$, updating steps of $\Theta$ on the perturbed dataset $k=10$, and a learning rate for updating spatial perturbations $\alpha' = 0.1/(\textrm{mean}(\nabla_{\bdelta} \sum_{r}||\hat\C(r;\Theta)) - \C(r;\{\x\})||_2))$, which normalizes the gradient. The total learning epochs are set to $m=2500$, the number of steps of updating $\Theta$ in a single epoch is $k=10$. It is important to note that, in accordance with NeRF training routines, we employ COLMAP \citep{schoenberger2016sfm} on the perturbed dataset to obtain camera positions during evaluation. All experiments were conducted on an NVIDIA A6000 GPU, with the time required for executing a poisoning attack on TensoRF for a single scene averaging around 6 hours.



\begin{table}[t]
\centering
\caption{Comparison of different perturbation strength $\rho$ of spatial deformation in poisoning TensoRF. We report PSNR/SSIM (higher is better) and LPIPS \citep{zhang2018unreasonable} (lower is better).}
\label{tab:perturb strength}
\begin{tabular}{ccccccc}
\toprule
          & \multicolumn{3}{c}{\textbf{Flowers}} & \multicolumn{3}{c}{\textbf{Orchids}} \\ \midrule
Strength  & PSNR$\uparrow$       & SSIM $\uparrow$      & LPIPS$\downarrow$      & PSNR $\uparrow$      & SSIM$\uparrow$       & LPIPS$\downarrow$      \\ \midrule
Original ($\rho=0$)  & 28.60      & 0.871      & 0.169      & 19.87      & 0.649      & 0.278      \\ 
$\rho=1$  & 24.54      & 0.739      & 0.392      & 18.92      & 0.535      & 0.452      \\ 
$\rho=2$  & 23.08      & 0.635      & 0.498      & 17.94      & 0.420      & 0.543      \\ 
$\rho=5$  & 21.15      & 0.525      & 0.574      & 17.11      & 0.330      & 0.577      \\ 
$\rho=10$ & 20.84      & 0.509      & 0.572      & 16.91      & 0.317      & 0.590      \\ 
$\rho=20$ & 18.95      & 0.462      & 0.594      & 14.98      & 0.291      & 0.606      \\ \bottomrule
\end{tabular}
\vspace{-0.1cm}
\end{table}

\begin{figure}
    \centering
    \includegraphics[height=8.5cm]{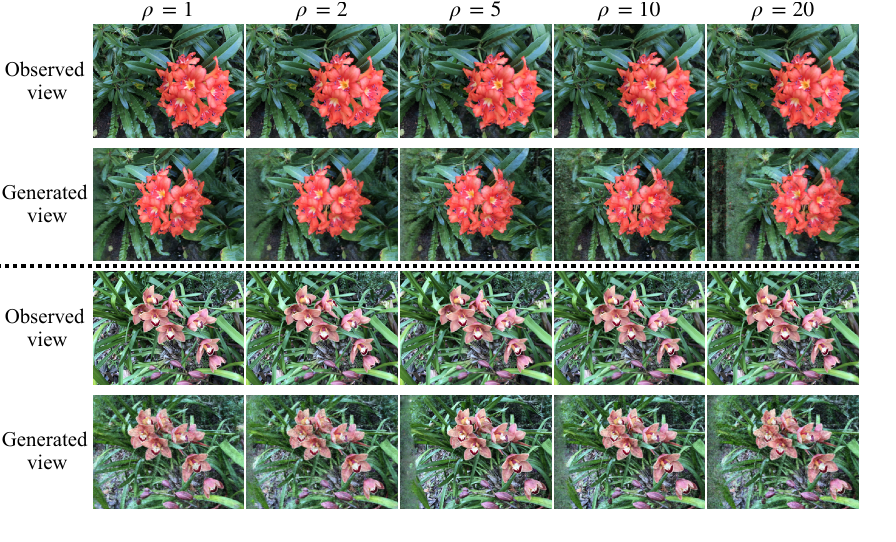}
    \vspace{-1cm}
    \caption{Comparisons on observed and generated views for the LLFF Flower and Orchids dataset with different poisoning strength $\rho$.}
    \label{fig:strength}
    \vspace{-0.35cm}
\end{figure}


\noindent\textbf{Results.} In \autoref{tab:general results}, our spatial poisoning attack demonstrates a significant reduction in the performance of TensoRF on both LLFF and Real-Forward-Facing datasets. A comparison of the novel views generated by the original observed views and the perturbed views can be found in \autoref{fig:resultall}. It becomes evident that there is a substantial difference between the images generated by the original and perturbed observed images, particularly at the edges of the novel views. The novel views generated from the perturbed images appear considerably more blurred compared to those generated from the original observed images. This observation underscores the effectiveness of our spatial poisoning attack in degrading the quality of the generated views, potentially compromising the utility of the NeRF models in various applications.

\begin{table}[]
\centering
\caption{The perturbed views generated by poisoning TensoRF can also reduce the performance of standard NeRF \citep{mildenhall2021nerf} and IBRNet \citep{wang2021ibrnet}. We report PSNR/SSIM (higher is better) and LPIPS \citep{zhang2018unreasonable} (lower is better). The labels "Ori" and "Poi" denote the performance of TensoRF when trained with clean and perturbed views, respectively. We present the averaged quantitative results on the eight LLFF scenes, with the red values in brackets indicating the reduction in performance..}
\label{tab:versatility}
\scalebox{0.93}{
\begin{tabular}{ccccccc}
\toprule
     & \multicolumn{3}{c}{\textbf{NeRF}\citep{mildenhall2021nerf}} & \multicolumn{3}{c}{\textbf{IBRNet}\citep{wang2021ibrnet}} \\
     \midrule
     & PSNR$\uparrow$ & SSIM$\uparrow$ & LPIPS$\downarrow$ & PSNR$\uparrow$ & SSIM$\uparrow$& LPIPS$\downarrow$\\
     \midrule
Ori. & 26.50     & 0.811     & 0.250     & 26.73      & 0.851      & 0.175     \\
Poi. & 16.39\red{(-10.11)}& 0.389\red{(-0.422)}& 0.661\red{(-0.411)}& 20.48\red{(-6.25)}& 0.577\red{(-0.274)}& 0.444\red{(-0.269)}\\
\bottomrule
\end{tabular}
}
\vspace{-0.4cm}
\end{table}

\subsection{Ablation studies}

\subsubsection{Different perturbation strength}
In this part, we investigate the impact of the spatial perturbation strength, denoted as $\rho$, on the rendering quality of TensoRF. We conduct experiments with $\rho$ values of 1, 2, 5, 10, and 20 on the LLFF Flower and Orchids scenes, maintaining the same settings as in previous experiments. The results are presented in \autoref{tab:perturb strength}. As the perturbation strength increases, the quantitative results of TensoRF generally decline, which aligns with the intuition that stronger perturbations can lead to more substantial degradation in the quality of the generated views. Additionally, we display the perturbed views and generated views in \autoref{fig:strength}, where it is evident that with a stronger perturbation, e.g., $\rho=20$, the generated views are markedly inferior compared to those with a milder perturbation ,e.g., $\rho=1$.

\vspace{-0.15cm}
\subsubsection{Adaptability}
In this part, we explore the adaptability of our poisoning attack (with $\rho=10$) on different NeRF backbones, including the original NeRF implementation \citep{mildenhall2021nerf} and another popular NeRF-based method IBRNet \citep{wang2021ibrnet}. We utilize the LLFF-Poisoned datasets generated in \autoref{tab:general results} to train the original NeRF model and finetune the IBRNet model. For original NeRF model, we adopt the same model settings as described in \citep{mildenhall2021nerf}, where a batch size of 4096 rays is used, each sampled at 64 coordinates in the coarse volume and 128 additional coordinates in the fine volume. We employ the Adam optimizer \citep{kingma2014adam} with a learning rate that starts at $5 \times 10^{-4}$ and decays exponentially to $5 \times 10^{-5}$ throughout the optimization process (other Adam hyperparameters remain at default values of $\beta_1 = 0.9$, $\beta_2 = 0.999$, and $\epsilon = 10^{-7}$). For IBRNet \citep{wang2021ibrnet}, we also use the same model and settings for fine tuning in \citep{wang2021ibrnet}, we use the Adam optimizer, the base learning rates for feature extraction network and IBRNet are $5\times 10^{-4}$ and $2 \times 10^{-4}$. We set a batch size of 4096 rays during finetuning.

In \autoref{tab:versatility} We present the quantitative results of the original NeRF model and IBRNet trained on the LLFF-Poisoned dataset, which was generated while poisoning the TensoRF model. The performance of the NeRF model is significantly reduced across all eight LLFF scenes, thereby validating the adaptability of our poisoning attacks. This outcome demonstrates that our proposed method can effectively compromise the performance of NeRF models even when applied to different variants, showcasing the versatility and potential impact of our poisoning technique on a broader range of NeRF-based applications.



\vspace{-0.25cm}
\section{Conclusion}
\vspace{-0.15cm}
\noindent\textbf{Limitation.}
Although our approach has demonstrated effectiveness in compromising the performance of NeRF models, there are several limitations that should be acknowledged. 1) The proposed poisoning attack algorithm involves a substantial amount of gradient calculations, which can be computational expensive. 2) Although our poisoning attack has been demonstrated to be adaptable across standard NeRF and IBRNet, its transferability to other types of 3D reconstruction models or deep learning architectures remains unexplored. 3) In our experiments, we have investigated the impact of different spatial perturbation strengths on the rendering quality of TensoRF. However, there might be other factors that influence the attack's effectiveness, such as the specific scene properties, which have not been thoroughly analyzed in our study.

In summation, our study marks a pioneering effort in safeguarding user privacy in NeRF models. This is achieved through the introduction of a subtle, imperceptible spatial deformation that disrupts NeRF training processes via poisoning attacks. Our proposed bi-level optimization algorithm, which incorporates a Projected Gradient Descent (PGD) based spatial deformation, has demonstrated its efficacy by successfully compromising the performance of NeRF models across real-image datasets. This research, therefore, acts as a crucial first step towards fortifying user privacy within NeRF models. Future work may focus on exploring alternative attack methods, as well as investigating potential defense mechanisms against such poisoning attacks to enhance the reliability and security of NeRF-based systems.

{\small
\bibliographystyle{unsrt}
\bibliography{main}
}

\clearpage
\appendix
\section{Discussion and Broader Impact}
NeRF's capacity to generate high-quality 3D content from readily available internet images is indeed a double-edged sword. The internet's vast repository of user-uploaded images opens up the possibility for unauthorized parties to generate 3D reconstructions of users' environments without consent, potentially inferring private and personally identifiable information. As we anticipate further enhancements in NeRF's ability to synthesize novel views, we are confronted with critical ethical dilemmas surrounding users' right to refuse the generation of such views from their images and the means to safeguard their privacy.

In light of these considerations, our study aims to integrate advancements in adversarial machine learning with the pressing issue of user privacy protection against NeRF's generative capabilities. We introduce a method to subtly modify 2D images that effectively disrupt view synthesis via NeRF, essentially acting as a "vaccine" for user-captured images against unauthorized 3D scene generation.

Our approach is rooted in exploiting NeRF's sensitivity to input variations, a vulnerability inherent in its optimization process. We employ spatial deformation to introduce such perturbations, effectively compromising NeRF's 3D space modeling. However, a significant challenge arises in executing these attacks during training time, considering NeRF's reliance on the provided images as training data.

The broader impact of our research extends beyond the immediate realm of NeRF and 3D scene generation. It touches upon the larger dialogue on privacy and ethical considerations in the use of generative machine learning models. It underscores the necessity for the research community to not only innovate but also to proactively address potential privacy and security implications of such advancements. Our work, therefore, serves as a crucial step towards ensuring responsible innovation in the field of machine learning and beyond.



\section{Additional results}
In this section, we report the per-scene result of \autoref{tab:general results} for 21 Real-Forward-Facing dataset and the per-scene result of \autoref{tab:versatility} for the LLFF-Poisoned datasets with standard NeRF and IBRNet.

\begin{table}[h]
\caption{Our poisoning algorithm significantly reduces the performance of TensoRF on the datasets of real images. We report PSNR/SSIM (higher is better) and LPIPS \citep{zhang2018unreasonable} (lower is better). The labels "Ori" and "Poi" denote the performance of TensoRF when trained with clean and perturbed views, respectively. Per-scene quantitative results for the 21 Real-Forward-Facing (full) scenes.}
\label{tab:perscenemain}
\scalebox{0.82}{
\begin{tabular}{ccccccccc}
\hline
                       &           & \textbf{colorfountain} & \textbf{colorspout} & \textbf{lumpyroots} & \textbf{redtoyota} & \textbf{succtrough} & \textbf{fenceflower} & \textbf{redspikey} \\ \hline
\multirow{2}{*}{PSNR$\uparrow$}  & Ori.  & 20.65                  & 26.62               & 18.76               & 22.00              & 24.56               & 20.59                & 18.60              \\
                       & Poi. & 19.20                  & 18.41               & 18.03               & 18.85              & 18.80               & 17.64                & 16.49              \\ \hline
\multirow{2}{*}{SSIM$\uparrow$}  & Ori.  & 0.5784                 & 0.8636              & 0.5623              & 0.690              & 0.803               & 0.649                & 0.580              \\
                       & Poi. & 0.316                  & 0.353               & 0.369               & 0.385              & 0.420               & 0.324                & 0.271              \\ \hline
\multirow{2}{*}{LPIPS$\downarrow$} & Ori.  & 0.324                  & 0.156               & 0.391               & 0.282              & 0.229               & 0.292                & 0.335              \\
                       & Poi. & 0.614                  & 0.632               & 0.575               & 0.618              & 0.584               & 0.559                & 0.593              \\ \hline
                       &           & \textbf{apples}        & \textbf{butcher}    & \textbf{livingroom} & \textbf{xmaschair} & \textbf{bikes}      & \textbf{leafscene}   & \textbf{lemontree} \\ \hline
\multirow{2}{*}{PSNR$\uparrow$}  & Ori.  & 24.42                  & 26.77               & 23.52               & 21.74              & 25.62               & 21.13                & 21.42              \\
                       & Poi.& 22.02                  & 22.68               & 21.94               & 18.09              & 18.83               & 15.47                & 18.84              \\ \hline
\multirow{2}{*}{SSIM$\uparrow$}  & Ori.  & 0.831                  & 0.865               & 0.764               & 0.702              & 0.828               & 0.742                & 0.669              \\
                       & Poi. & 0.670                  & 0.697               & 0.681               & 0.419              & 0.422               & 0.288                & 0.325              \\ \hline
\multirow{2}{*}{LPIPS$\downarrow$} & Ori.  & 0.201                  & 0.330               & 0.313               & 0.300              & 0.176               & 0.222                & 0.256              \\
                       & Poi. & 0.396                  & 0.509               & 0.458               & 0.592              & 0.587               & 0.592                & 0.559              \\ \hline
                       &           & \textbf{magnolia}      & \textbf{piano}      & \textbf{redplant}   & \textbf{stopsign}  & \textbf{skeleton}   & \textbf{house}       & \textbf{pond}      \\ \hline
\multirow{2}{*}{PSNR$\uparrow$}  & Ori.  & 16.55                  & 24.23               & 19.87               & 22.76              & 22.12               & 22.29                & 21.99              \\
                       & Poi. & 14.38                  & 19.96               & 15.61               & 17.66              & 20.94               & 19.12                & 18.75              \\ \hline
\multirow{2}{*}{SSIM$\uparrow$}  & Ori.  & 0.542                  & 0.790               & 0.655               & 0.734              & 0.716               & 0.740                & 0.644              \\
                       & Poi. & 0.189                  & 0.587               & 0.243               & 0.369              & 0.667               & 0.428                & 0.317              \\ \hline
\multirow{2}{*}{LPIPS$\downarrow$} & Ori.  & 0.365                  & 0.274               & 0.289               & 0.223              & 0.419               & 0.200                & 0.305              \\
                       & Poi. & 0.667                  & 0.549               & 0.654               & 0.515              & 0.576               & 0.502                & 0.591              \\ \hline
\end{tabular}
}

\end{table}

\begin{table}[h]
\centering
\caption{The perturbed views from the LLFF-Poisoned dataset can also reduce the performance of standard NeRF \citep{mildenhall2021nerf} and IBRNet \citep{wang2021ibrnet}. We report PSNR/SSIM (higher is better) and LPIPS \citep{zhang2018unreasonable} (lower is better). The labels "Ori" and "Poi" denote the performance when trained with clean and perturbed views, respectively. We present per-scene and averaged quantitative results on the eight LLFF scenes (since IBRNet didn't provide per-scene results, we use 'NA' refers the missing result).}
\label{tab:persceneversatility}
\scalebox{0.93}{
\begin{tabular}{cccccccccc|c}
\toprule
 \multicolumn{11}{c}{\textbf{NeRF}\citep{mildenhall2021nerf}} \\
 \hline
                       &           & \textbf{Fern} & \textbf{Fortress} & \textbf{Leaves} & \textbf{Flower} & \textbf{Horns} & \textbf{Orchids} & \textbf{Room} & \textbf{T-Rex} & \textbf{AVG} \\ \midrule
\multirow{2}{*}{PSNR$\uparrow$}  & Ori.  & 25.17         & 31.16             & 20.92           & 27.40           & 27.45          & 20.36            & 25.17         & 26.80          & 26.50        \\
                       & Poi. & 15.38         & 18.34             & 12.29           & 16.14           & 17.48          & 13.68            & 20.12         & 17.71          & 16.39        \\ \midrule
\multirow{2}{*}{SSIM$\uparrow$}  & Ori.  & 0.792         & 0.881             & 0.690           & 0.827           & 0.828          & 0.641            & 0.948         & 0.880          & 0.811        \\
                       & Poi. & 0.384         & 0.511             & 0.163           & 0.374           & 0.420          & 0.160            & 0.669         & 0.430          & 0.389        \\ \midrule
\multirow{2}{*}{LPIPS$\downarrow$} & Ori.  & 0.280         & 0.171             & 0.316           & 0.219           & 0.268          & 0.321            & 0.178         & 0.249          & 0.250        \\
                       & Poi. & 0.667         & 0.614             & 0.685           & 0.668           & 0.679          & 0.730            & 0.584         & 0.658          & 0.661        \\ \bottomrule
\end{tabular}
}

\vspace{0.3cm}

\scalebox{0.93}{
\begin{tabular}{cccccccccc|c}
\toprule
 \multicolumn{11}{c}{\textbf{IBRNet}\citep{wang2021ibrnet}} \\
 \midrule
                       &           & \textbf{Fern} & \textbf{Fortress} & \textbf{Leaves} & \textbf{Flower} & \textbf{Horns} & \textbf{Orchids} & \textbf{Room} & \textbf{T-Rex} & \textbf{AVG} \\ \midrule
\multirow{2}{*}{PSNR$\uparrow$}  & Ori.  & NA             & NA                 & NA               & NA               & NA              & NA                & NA             & NA              & 26.73        \\
                       & Poi. & 20.18         & 26.83             & 15.61           & 20.64           & 21.13          & 16.91            & 21.96         & 20.64          & 20.48        \\ \midrule
\multirow{2}{*}{SSIM$\uparrow$}  & Ori.  & NA             & NA                 & NA               & NA               & NA              & NA                & NA             & NA              & 0.851        \\
                       & Poi. & 0.537        & 0.723            & 0.397          & 0.543          & 0.614         & 0.375           & 0.785        & 0.635         & 0.576       \\ \midrule
\multirow{2}{*}{LPIPS$\downarrow$} & Ori.  & NA             & NA                 & NA               & NA               & NA              & NA                & NA             & NA              & 0.175        \\
                       & Poi. & 0.495        & 0.326            & 0.535          & 0.480          & 0.435         & 0.519           & 0.356        & 0.406         & 0.444       \\ \bottomrule
\end{tabular}
}
\end{table}
\clearpage
\begin{figure}
    \centering
    \includegraphics[height=22cm]{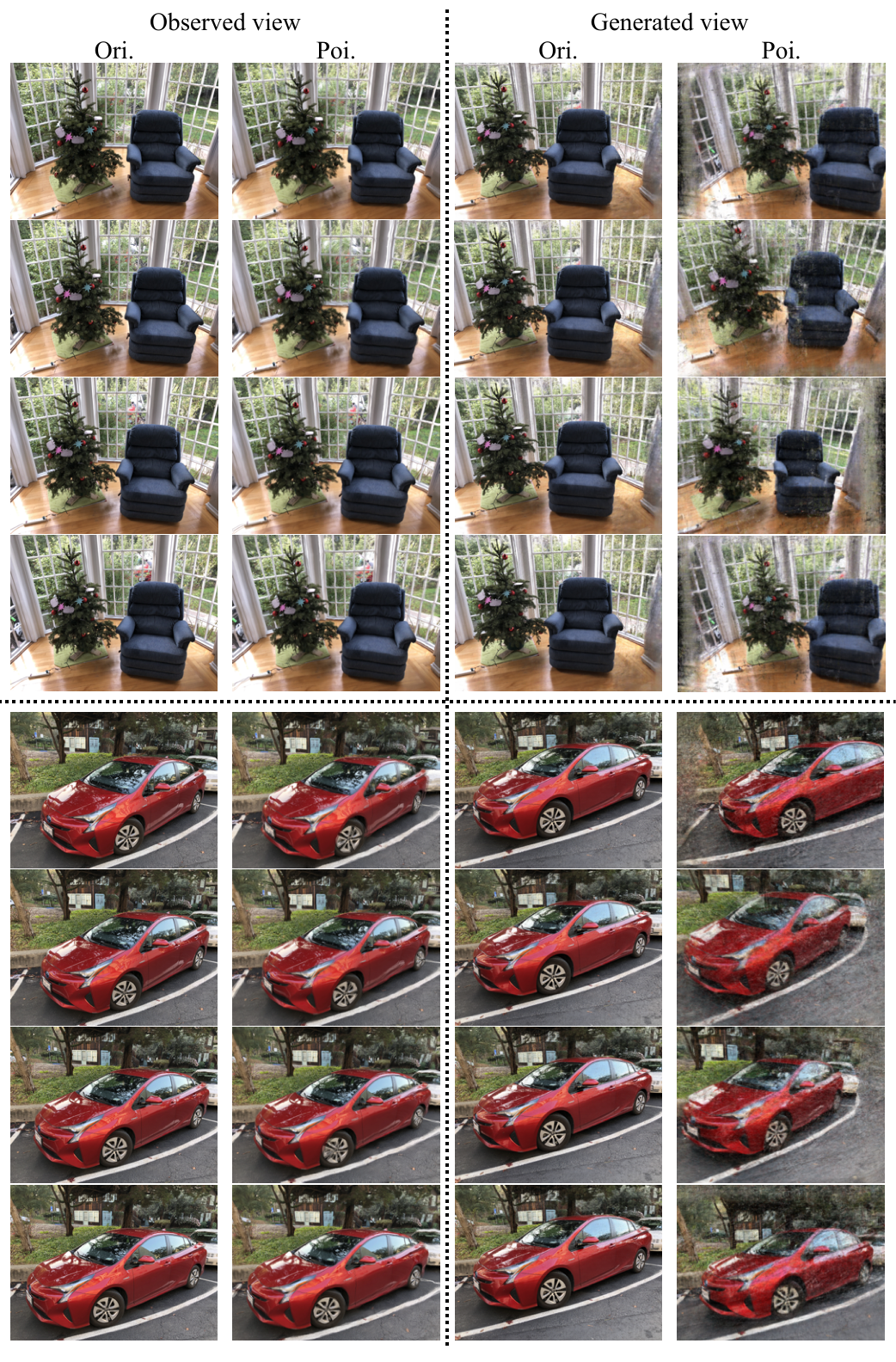}
    \vspace{-0.8cm}
    \caption{Comparisons on observed and generated views for the Real-Forward-Facing dataset. For the poisoning attack we apply spatial deformation with strength $\rho=10$. It is evident that the generated views are significantly impacted (particularly at the edges of the images) by subtle alterations in the observed views.}
    \label{fig:add1}
\end{figure}

\begin{figure}
    \centering
    \includegraphics[height=22cm]{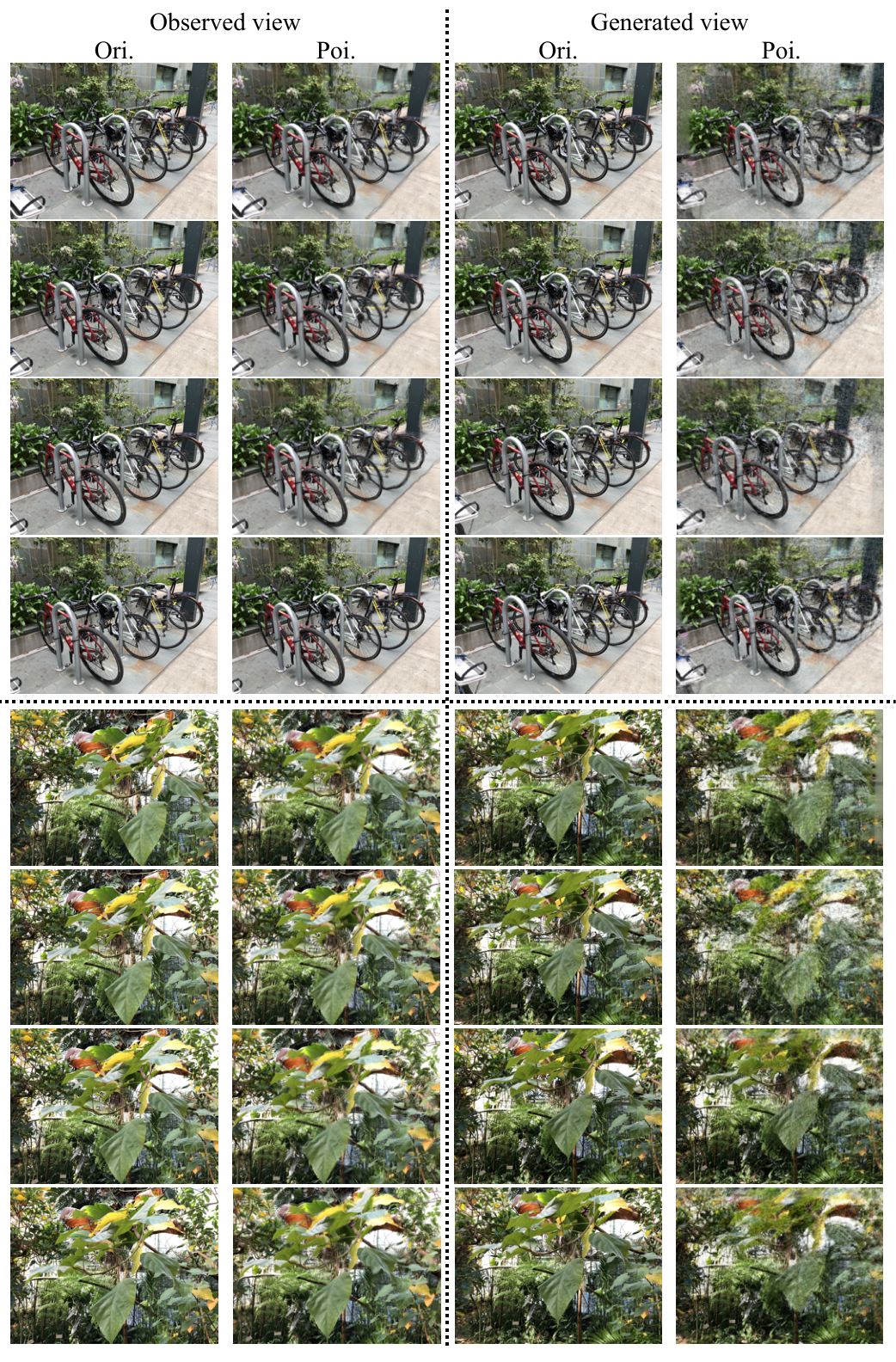}
    \vspace{-0.8cm}
    \caption{Comparisons on observed and generated views for the Real-Forward-Facing dataset. For the poisoning attack we apply spatial deformation with strength $\rho=10$. It is evident that the generated views are significantly impacted (particularly at the edges of the images) by subtle alterations in the observed views.}
    \label{fig:add2}
\end{figure}

\begin{figure}
    \centering
    \includegraphics[height=22cm]{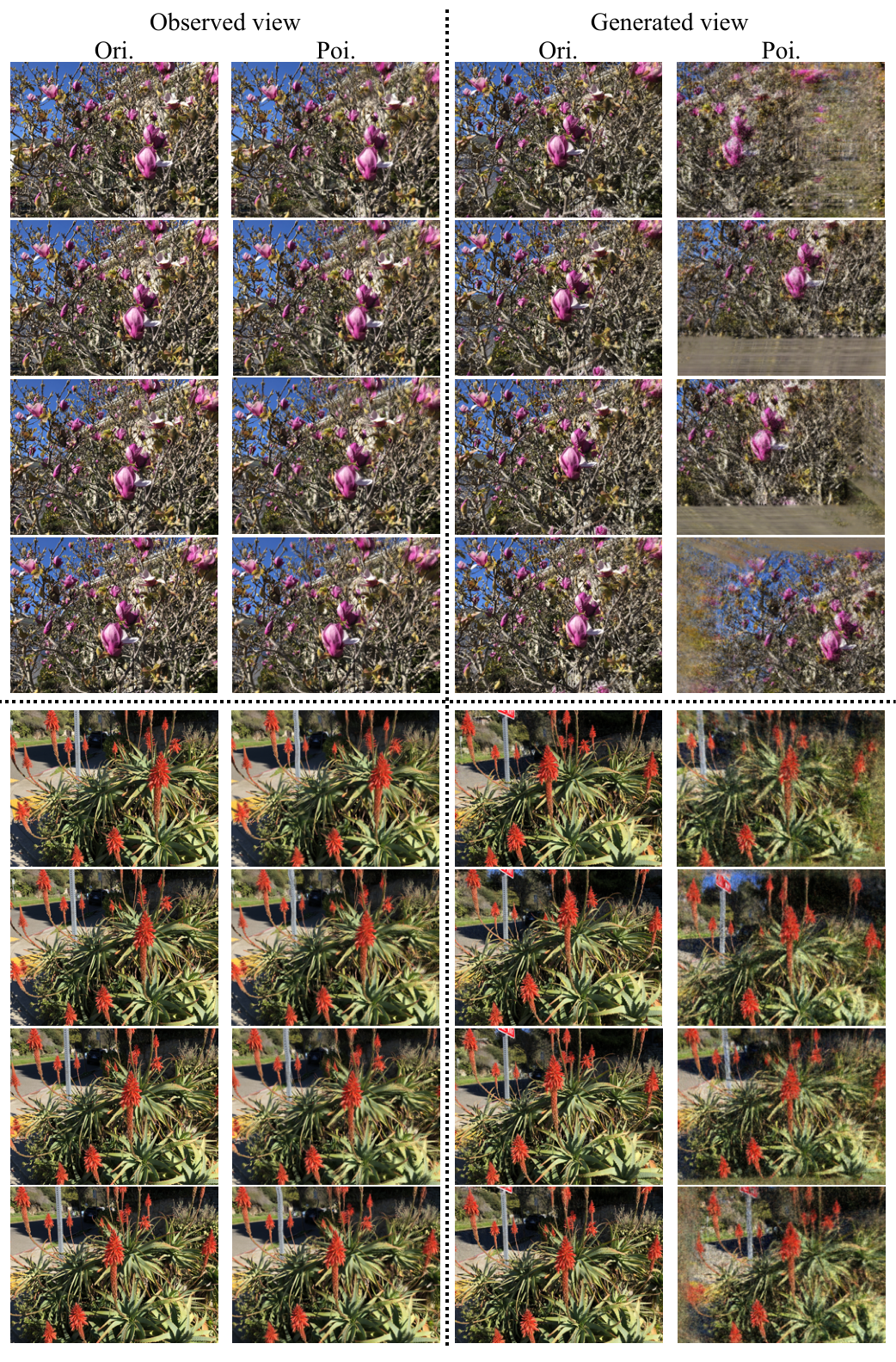}
    \vspace{-0.8cm}
    \caption{Comparisons on observed and generated views for the Real-Forward-Facing dataset. For the poisoning attack we apply spatial deformation with strength $\rho=10$. It is evident that the generated views are significantly impacted (particularly at the edges of the images) by subtle alterations in the observed views.}
    \label{fig:add3}
\end{figure}

\end{document}